\documentclass[
]{ceurart}

\usepackage{listings}
\usepackage{cleveref}
\usepackage{hyperref}
\begin{document}

\copyrightyear{2026}
\copyrightclause{Copyright for this paper by its authors.
  Use permitted under Creative Commons License Attribution 4.0
  International (CC BY 4.0).}

\conference{CLEF 2026: Conference and Labs of the Evaluation Forum, September 21–24, 2026, Jena, Germany}

\title{Team DArgk at the 2026 ELOQUENT lab for evaluating generative language model quality:\\Residuals of Humanity: AI Detection Evasion via GRPO Fine-Tuning}

\title[mode=sub]{Notebook for the Voight-Kampff Task on Eloquent Lab 2026 at CLEF 2026}

\author[1,2]{Antonela Tommasel}[%
orcid=0000-0001-6091-8305,
email={antonela.tommasel@isistan.unicen.edu.ar, antonela.tommasel@jku.at},
]
\cormark[1]
\fnmark[1]
\address[1]{Johannes Kepler University, Linz, Austria}
\address[2]{CONICET - UNICEN, Tandil, Buenos Aires, Argentina}

\author[3,2]{Juan Manuel Rodriguez}[%
orcid=0000-0002-1130-8065,
email=jmro@cs.aau.dk,
]
\cormark[1]
\fnmark[1]
\address[3]{Department of Computer Science, Aalborg University, Aalborg, Denmark}

\cortext[1]{Corresponding author.}
\fntext[1]{These authors contributed equally.}

\begin{abstract}
Large language models (LLMs) can generate fluent and coherent text that is increasingly difficult to distinguish from human writing, motivating the development of automatic AI-generated text detectors. However, the robustness of such detectors under adversarial generation remains uncertain. This paper presents \texttt{SHADE} (\textit{Stochastic Human-like generation via Adversarial Detector Evasion}), a reinforcement learning framework that formulates detector evasion as a policy optimization problem. Instead of applying post-hoc perturbations or prompting-based rewriting, \texttt{SHADE} fine-tunes an instruction-tuned LLaMA model with Group Relative Policy Optimization (GRPO), using feedback from a surrogate detector based on the PAN 2025 \textit{mdok} system. Our experiments show that full fine-tuning with a small KL regularization penalty achieves $98.5\%$ surrogate evasion, compared to $1.5\%$ for the base model, while LoRA-based adaptation is substantially less effective under regularization. Linguistic analysis reveals that successful evasion is associated with shorter, simpler, and less lexically diverse outputs, suggesting that high detector evasion does not necessarily correspond to more human-like writing. In the official Voight-Kampff competition setting, our submissions ranked sixth and seventh, indicating that optimization against a single surrogate detector only partially transfers to unseen evaluation classifiers. These results highlight both the potential and limitations of reinforcement learning for adversarial AI-text generation and motivate more robust, multi-detector evaluation protocols for AI-generated text detection.
\end{abstract}


\maketitle

\section{Introduction}
The rapid advances in Large Language Models (LLMs), which are now capable of producing fluent, coherent, and human-like text across a wide range of domains and writing styles, have made it increasingly difficult to determine whether a text is written by a human or generated by an AI system \cite{YANG2025126694}. This distinction is becoming important in settings where authorship, originality, accountability and trust are central, including education, journalism, scientific communication, online moderation and misinformation detection \cite{detectRL}. 
As LLM-generated content becomes easier to produce, adapt and distribute at scale, automatic AI-generated text detection has emerged as a key mechanism for identifying synthetic text and supporting decisions about content provenance. 
A growing body of work has therefore proposed detection methods based on perplexity, stylometric features, watermarking signals or fine-tuned transformers \cite{perplexity, AdaDetectGPT, OUTFOX, detectRL}. However, the reliability of these detectors remains uncertain, particularly when generated texts are intentionally modified to appear less machine-generated. Their robustness under adversarial conditions is thus still an open question \cite{ICLR2025_attack, OUTFOX}.

Motivated by this challenge, the Voight-Kampff task~\cite{eloquenttask2026voightkampff} at the ELOQUENT 2026 lab~\cite{eloquent2026overview} asks participants to generate text that evades unseen AI detectors submitted by participants of PAN 2026. In the previous edition of the Voight-Kampff task, the strongest approaches relied on text perturbation and translation-based transformations, whereas persona-based and stylistic rewriting strategies (i.e., prompting-based methods) achieved weaker results \cite{clef25creo, clef2025poojan}. 
Data-centric approaches have also explored fine-tuning with LoRA on human-written corpora to better mimic the distribution of human texts in the training data \cite{clef2025gunti}. However, these methods largely treat evasion as either a post-processing problem or a static alignment problem, rather than as a dynamic optimization objective guided by detection feedback.
 
In this paper, we propose \texttt{SHADE} (\textit{Stochastic Human-like generation via Adversarial Detector Evasion}), a framework that formulates detector evasion as a reinforcement learning problem. Instead of manipulating outputs post-hoc or curating training data to approximate human style, \texttt{SHADE} directly optimizes a language model's generation policy to reduce the confidence of an adversarial surrogate detector. 
Concretely, we fine-tune LLaMA 3.2 3B instruct using Group Relative Policy Optimization (GRPO), with rewards derived from a surrogate detector based on last year's Voight-Kampff task, namely \textit{mdok} \cite{mdok}. This adversarial formulation encourages the model to internalize evasion-oriented generation strategies, rather than relying exclusively on surface-level perturbations or prompting-based transformations.

The contributions of this paper are threefold:
(i) we formulate AI-generated text detector evasion as a reinforcement learning problem driven by adversarial detector feedback;
(ii) we introduce \texttt{SHADE}, a GRPO-based fine-tuning framework that optimizes an LLM against a surrogate from the previous Voight-Kampff task; 
and (iii) we empirically analyze the relationship between detector evasion and textual properties, showing that high evasion rates can emerge through linguistic simplification rather than through more elaborate human-style generation.

Our experiments show that full fine-tuning with a small KL regularization penalty achieves a $98.5\%$ evasion rate against the surrogate detector, compared to only $1.5\%$ for the base LLM. To better understand this behavior, we analyze the trade-offs between evasion and textual characteristics such as repetition, lexical richness, and readability. This analysis reveals that the model primarily evades detection by simplifying its language, suggesting that successful detector evasion does not necessarily correspond to richer or more human-like writing. We further show that parameter efficient tuning with LoRA is substantially less effective in this adversarial reinforcement learning setting, motivating full model adaptation for the proposed task.
 
The remainder of this paper is organized as follows. \Cref{sec:rel_work} reviews related work on AI detection evasion. 
\Cref{sec:SHADE} formalizes the problem and details \texttt{SHADE}, including its training objective, surrogate detector, and inference procedure. 
\Cref{sec:model_analysis} analyzes the fine-tuning configurations and the linguistic properties of the generated texts. 
\Cref{sec:final_results} reports results on the official Voight-Kampff task evaluation. 
\Cref{sec:conclusions} concludes the paper and \Cref{sec:artifacts} links the artifacts related to this work. 

\section{Related work\label{sec:rel_work}}

Research on AI-generated text detection and evasion lies at the intersection of two closely related questions: how such detectors can be attacked, and which evasion strategies have proven effective in shared-task settings. 
We therefore first discuss \textit{AI-generated text detection}, focusing on statistical, supervised and watermarking-based approaches for distinguishing machine-generated from human-written text. We then review \textit{attacks on AI-generated text detectors}, distinguishing input-level transformations from model-level adaptations. Finally, we discuss \textit{evasion approaches in the Voight-Kampff task 2025}, which provides the most direct empirical context for our contribution. This structure allows us to position \texttt{SHADE} with respect to both the broader literature and the specific shared-task setting.

\subsubsection*{AI-generated Text Detection}

Recent surveys characterize AI-generated text detection as a rapidly evolving field that spans several methodological families. \citet{YANG2025126694} organize existing approaches around three main challenges: improving classifier training, exploiting intrinsic attributes of language models and embedding information such as watermarks into generated text. They further distinguish black-box detectors, which rely only on input--output behavior and observable textual characteristics, from white-box detectors, which assume access to internal model information such as architecture, parameters, logits, or generation procedures. This distinction is particularly relevant for evasion-oriented settings, where the attacker may only have access to detector outputs or surrogate feedback rather than to the target detector itself. This broader taxonomy also helps explain why detector evasion remains difficult to address with a single class of defenses. Training-based detectors depend on the representativeness of labeled data, intrinsic-attribute methods often rely on access to model probabilities or representations, and watermarking approaches require some form of information embedding during generation. As a result, detection performance can vary substantially depending on model access, computational assumptions, domains, and attack conditions.

The task of distinguishing machine-generated text from human-written text has been mainly approached from two broad directions: zero-shot statistical methods and supervised classification. 
Statistical detectors exploit regularities in the probability distribution of LLM-generated text, such as the tendency of generated text to exhibit lower entropy or higher likelihood under a reference model. 
Binoculars~\cite{perplexity}, for instance, performs classification by comparing perplexity-based scores from two language models that share a tokenizer, enabling zero-shot detection without labeled training data. 
AdaDetectGPT~\cite{AdaDetectGPT} extends perturbation-based detection with statistical guarantees, using adaptive thresholds to control false-positive rates across domains. 
Similarly, \citet{seeliger2025human} proposed a statistically grounded approach for PAN 2025 \cite{pan2025}. Their method first computes world-level correlations from a binary-term document matrix and the associated human/AI labels, then maps each text into a sequence of correlation values interpreted as a signal. These correlation signals, combined with hand-crafted statistical features, provide a lightweight and interpretable alternative to detectors based on LLM backbones.

A second line of work frames AI-generated text detection as a supervised binary classification problem~\cite{detectRL,OUTFOX,mdok}. 
DetectRL~\cite{detectRL} provides a benchmark for evaluating such detectors under more realistic conditions, including paraphrase attacks and domain shifts, and shows that many classifiers degrade substantially when exposed to adversarially modified inputs. 
OUTFOX~\cite{OUTFOX} addresses this limitation by augmenting the training set with adversarially generated essays, i.e., texts explicitly generated to challenge the detector. 
In the PAN 2025 setting, mdok~\cite{mdok} ranked among the top-performing systems, combining an LLM backbone with robust fine-tuning to achieve strong performance in both binary and multiclass detection. 
The continued development of detection benchmarks at PAN 2026~\cite{pan2026} reflects the rapid evolution of the field. However, it also highlights an unresolved problem. As detectors become stronger, evasion methods may need to become adaptive rather than relying on fixed transformations or manually designed prompts.

\subsubsection*{Attacks on AI-generated Text Detectors} 

The problem of evading AI-generated text detectors can be situated withing the broader paradigm of adversarial machine learning~\cite{adversarial}, where an attacker modifies inputs of model behavior to induce misclassification. In the context of text detection, a common strategy is to apply input-level transformations to generated text. These transformations include character-level perturbations, synonym substitutions, paraphrasing and translation-based rewriting, with the goal of moving the text away from the distribution recognized by the detector \cite{clef25creo}. Such methods can be effective because they alter superficial or statistical cues used by detectors. However, they may also introduce artifacts, reduce readability or produce unnatural phrasing. 
A recent survey \cite{YANG2025126694} similarly identifies adversarially generated text as an emerging challenge for AI-generated text detection, noting that such texts can deceive existing detectors and increase false-negative rates. This supports the need for evaluation settings in which detectors are tested not only on standard generated text, but also against adaptive or adversarial generation strategies.

A complementary strategy adapts the source model itself. Instead of modifying generated outputs after generation, model-level approaches fine-tune the generator so that its output distribution shifts toward texts that are less likely to be classified as machine-generated. \citet{ICLR2025_attack} propose HUMPA, a proxy-attack strategy that fine-tunes a small surrogate model using Direct Preference Optimization (DPO) \cite{dpo}. Their approach reduces detector performance substantially, achieving an average AUROC drop of $70.4\%$ across multiple datasets. Importantly, this gain comes with a trade-off. more aggressive fine-tuning can reduce generation quality, making the regularization parameter $\beta$ central for balancing evasion performance and text quality \cite{ICLR2025_attack}. This finding is particular relevant to our work, as \texttt{SHADE} also optimizes a generator against detector feedback, but does so through reinforcement learning rather than preference optimization. 

\subsubsection*{Evasion Approaches at Voight-Kampff Task 2025} 

The 2025 edition of the Voight-Kampff task~\cite{pan2025} produced a range of evasion strategies that illustrate both the promise and the limitations of existing approaches. \citet{clef25creo} evaluated four methods: Text Perturbation, Lost in Translation, Persona Immersion, and ADHD Writing. Text Perturbation applies character-level modifications to obscure the generated text. Lost in Translation translates the text into another language and then back into the original language, for example from English to Spanish and back to English, with the goal of shifting the distribution of the generated text. Persona Immersion and ADHD Writing, in contrast, are prompting-based methods that ask the LLM to imitate human writing either directly or through a specific cognitive or stylistic profile. The authors found that Lost in Translation achieved the best overall evasion performance, followed by Text Perturbation and ADHD Writing. These results suggest that translation-based rewriting can be more effective than direct prompt-based human imitation, likely because translation changes lexical and syntactic patterns more substantially than style instructions alone. At the same time, these methods remain post-processing strategies: they transform a generated text after it has been produced, rather than modifying the generation process itself.

\citet{clef2025poojan} explored a related translation-based strategy using Hindi-to-Spanish re-translation. Although this approach is conceptually similar to Lost in Translation, the resulting texts exhibited awkward phrasing, repetition, grammatical errors, inconsistent idioms, and uneven sentence structure. These results suggest that translation-based evasion may succeed partly by introducing distributional shifts that detectors fail to capture, but that these shifts can come at the cost of fluency and textual quality. 
Finally, \citet{clef2025gunti} proposed fine-tuning LLaMA 3 with LoRA on cleaned human-written corpora aiming to shift the model distribution toward human-written text. This data-centric strategy can preserve output quality because the model is trained on human-authored examples. However, it treats evasion as a static alignment problem, the model is optimized to resemble a human-text corpus rather than to adapt directly to detector feedback. Moreover, as it relies on the original PAN dataset as part of the training data, its effectiveness may depend on the match between the training distribution and the target evaluation setting. 

In contrast, \texttt{SHADE} frames detector evasion as a dynamic optimization problem. Like model-level adaptation approaches, it fine-tunes the generator rather than perturbing outputs post hoc. However, unlike LoRA-based alignment to human corpora \cite{clef2025gunti}, \texttt{SHADE} does not require a corpus of human-written texts. Its objective is not to imitate existing human writing directly, but to optimize generation behavior against an adversarial surrogate detector. Moreover, unlike DPO-based proxy attacks \cite{ICLR2025_attack}, which optimize a generator offline using a fixed set of paired preferences, \texttt{SHADE} uses online reinforcement learning with detector confidence as a per-sample reward signal. Specifically, we adopt GRPO~\cite{grpo} and define the reward as the complement of the confidence score assigned by a static, pretrained adversarial detector. This places \texttt{SHADE} within the broader paradigm of Reinforcement Learning from AI Feedback, but in a setting where the feedback is used to optimize detector evasion. In this sense, our contribution is to apply feedback-driven policy optimization to a problem that has more commonly been addressed through post-hoc transformations or static distributional alignment.

\section{\texttt{SHADE}: \textit{Stochastic Human-like generation via Adversarial Detector Evasion}\label{sec:SHADE}}

The goal of the Voight-Kampff task is to evaluate whether text generated by a language model can be made difficult to distinguish from human-written text. In the official evaluation setting, generated texts are assessed by unseen AI-detection classifiers. Since participants do not have access to these target detectors, the task can be viewed as a black-box evasion problem in which the generator must produce prompt-consistent texts that are unlikely to be classified as AI-generated. %In this section, we first formalize this problem in \Cref{sec:problem_def}, and then describe our proposed solution in \Cref{sec:sol}.

\subsection{Problem definition\label{sec:problem_def}}

Let $\mathcal{P}$ denote the prompt space and $\mathcal{Y}$ the space of possible text outputs. We define a parameterized language model generator $\pi_\theta : \mathcal{P} \rightarrow \mathcal{Y}$, which induces a distribution over outputs $y \in \mathcal{Y}$ for a given prompt $p \in \mathcal{P}$. Let $D_\phi : \mathcal{Y} \rightarrow [0,1]$ be an unknown AI-detection classifier, where $D_\phi(y)$ denotes the probability that text $y$ is classified as AI-generated.
The objective is to learn parameters $\theta$ such that the generated texts satisfy the semantic and stylistic constraints of the input prompt while minimizing the detector confidence. Formally, we define the evasion objective as:
\begin{equation}
   \theta^* = \arg\max_{\theta} \mathbb{E}_{p \sim \mathcal{P}} \left[ \mathbb{E}_{y \sim pi_\theta(p)} \left[ 1 - D_\phi(y) \right] \right] 
\end{equation}

Since $D_\phi$ is not accessible during training, we approximate this objective using a surrogate detector $\hat{D}_\phi$. The resulting optimization problem therefore seeks to adapt the generation policy so that its outputs receive low AI-generated probability scores under the surrogate, while remaining close enough to the original model to preserve generation quality.

\subsection{Solution\label{sec:sol}}

We propose \texttt{SHADE}, \textit{Stochastic Human-like generation via Adversarial Detector Evasion}, a framework that combines adversarial learning \cite{adversarial} and reinforcement learning to adapt a base language model for detector evasion. The central idea is to modify the weights of the base model so that evasion-oriented generation behavior is internalized by the model, rather than applied through post-hoc perturbations, translation, or manually designed prompts. To train the model, we use the datasets from the Voight-Kampff tasks of Eloquent 2024, 2025, and 2026, together with the prompt instructions described in \Cref{sec:prompt}.

As the official AI-detection classifiers $D_\phi$ are not available, we train against a surrogate detector. In particular, we use a modified version of \textit{mdok} \cite{mdok}, one of the best-performing classifiers in PAN 2025. While the original \textit{mdok} system uses \texttt{Qwen3-14BBase} \cite{yang2025qwen3}, our surrogate replaces this backbone with \texttt{FacebookAI/roberta-base}~\cite{roberta} due to GPU memory constraints. We refer to this surrogate as \textit{mdok-}\texttt{roberta}. To approximate the behavior of the original mdok detector, we trained \textit{mdok-}\texttt{roberta} on the PAN 2025 dataset. On the PAN 2025 validation set, \textit{mdok-}\texttt{roberta} achieved an F1 score of $99\%$. We therefore use it as an adversarial oracle, assuming that detectors used in PAN 2026 \cite{pan2026} will exploit similar distributional regularities.

Given this surrogate detector, we fine-tune \texttt{LLaMA 3.2 3B} in its instruction-tuned version\footnote{\href{https://huggingface.co/meta-llama/Llama-3.2-3B-Instruct}{https://huggingface.co/meta-llama/Llama-3.2-3B-Instruct}}~\cite{grattafiori2024llama3herdmodels}.
We use Group Relative Policy Optimization (GRPO)~\cite{grpo} as the optimization algorithm. In this setting, $\pi_\theta$ denotes the model being optimized, $\pi_\text{ref}$ denotes the original base model before reinforcement learning, and $\hat{D}_\phi(y)$ denotes the surrogate detector score. Since $\hat{D}_\phi(y)$ estimates the probability that $y$ is AI-generated, we define the reward as $r(y)=1-\hat{D}_\phi(y)$. Thus, outputs that receive lower AI-generated probability scores obtain higher rewards.

Given a training batch, let $r_i=r(y_i)$ be the reward assigned to the $i$-th generated text, and let $\text{mean}(\mathbf{r})$ and $\text{std}(\mathbf{r})$ denote the mean and the standard deviation of rewards across the current batch. Following the GRPO formulation, the advantage assigned to each token $t$ in the $i$-th outputs is computed by normalizing the reward within the current group:

\begin{equation}
    \hat{A}_{i,t} = \frac{r_i - \text{mean}(\mathbf{r})}{\text{std}(\mathbf{r})}
\end{equation}

Then, the loss function is defined as:

\begin{equation}
\mathcal{L}_{\mathrm{GRPO}}(\theta)
=
-
\frac{1}{\sum_{i=1}^{G} |\mathcal{O}_i|}
\sum_{i=1}^{G}
\sum_{t=1}^{|\mathcal{O}_i|}
\left[
\frac{
\pi_{\theta}(o_{i,t} \mid q, o_{i,<t})
}{
\left[
\pi_{\theta}(o_{i,t} \mid q, o_{i,<t})
\right]_{\mathrm{old}}
}
\hat{A}_{i,t}
-
\beta \mathbb{D}_{\text{KL}}
\left[
\pi_{\theta} \,\|\, \pi_{\mathrm{ref}}
\right]
\right]    
\end{equation}

This formulation follows the implementation provided by the \textit{Hugging Face TRL} library\footnote{\href{https://huggingface.co/docs/trl/grpo_trainer}{https://huggingface.co/docs/trl/grpo\_trainer}}, which slightly differs from the original GRPO objective. 
For the KL-divergence, we used the same unbiased estimator as in the original implementation\footnote{\href{http://joschu.net/blog/kl-approx.html}{http://joschu.net/blog/kl-approx.html}}:

\begin{equation}
    \mathbb{D}_{\text{KL}}\left[\pi_\theta \|\pi_{\text{ref}}\right] = \frac{\pi_{\text{ref}}(o_{i,t} \mid q, o_{i,<t})}{\pi_\theta(o_{i,t} \mid q, o_{i,<t})}-\log (\frac{\pi_{\text{ref}}(o_{i,t} \mid q, o_{i,<t})}{\pi_\theta(o_{i,t} \mid q, o_{i,<t})})-1
\end{equation}

Although the KL-divergence term is usually omitted when training models using GRPO~\cite{NEURIPS2025_ed873d79}, we found that using a small $\beta$ improves the textual properties of the generated outputs. We analyze this effect in \Cref{sec:model_analysis}.

\subsection{Prompt modification\label{sec:prompt}}

For each input instance, the Voight-Kampff task provides a base prompt that specifies the text-generation request. We adapt this task-provided prompt to the chat format expected by the instruction-tuned \texttt{LLaMA 3.2 3B} model. Specifically, we wrap the original prompt in a chat template and add system-level instructions intended to exploit the model's instruction following capabilities\footnote{\href{https://huggingface.co/docs/transformers/en/chat_templating}{https://huggingface.co/docs/transformers/en/chat\_templating}}:

\begin{verbatim}
    [
        {"role": "system", 
         "content":  "You are a helpful assistant that generates helpful 
                      answers. You will avoid pleasantries and small talk, 
                      focusing on the task at hand."},
        
        {"role": "system", 
         "content": "You will avoid short paragraphs and bullet points."},
        
        {"role": "user", 
         "content": BASE_PROMPT + "\n\nThis is the text"},
        
        {"role": "assistant", 
         "content": ""},
    ]
\end{verbatim}

This formatting allows the prompt to match the conversational structure expected by the instruction-tuned model while preserving the original task instructions.

\subsection{Final text generation}

At inference time, text generation is performed using the chat-based template described in \Cref{sec:prompt}. We use stochastic decoding to introduce controlled variation across candidate outputs. Sampling is controlled using temperature and top-$p$ sampling. Temperature adjusts the randomness of the output distribution, while top-$p$ sampling restricts generation to the smallest set of tokens whose cumulative probability exceeds the specified threshold. 
Generation continues until either an end-of-sequence condition is reached or the maximum number of new tokens has been produced.
We set the temperature to $0.8$, the top-$p$ to $0.9$, and the maximum number of new tokens to $1000$. 

For each prompt, we generate multiple candidate texts. Each candidate is evaluated using the surrogate detector, which assigns a score in $[0,1]$ estimating the probability that the text was generated by an AI system. Lower scores indicate more human-like writing. The system iteratively generates candidates until either a text with an AI-generated probability below the acceptance threshold is found or the maximum number of attempts is reached. We set the acceptance threshold to $0.2$ and the maximum number of attempts to $10$. If no candidate satisfies the threshold, the candidate with the lowest surrogate detector score is selected as the final output.

\section{Model analysis\label{sec:model_analysis}}

When training \texttt{SHADE}, we evaluated several fine-tuning configurations. 
First, we compared parameter-efficient fine-tuning with LoRA against full fine-tuning, where all model weights are updated. 
Second, we varied the KL regularization coefficient $\beta$, which controls the penalty for deviating from the reference policy, using values $\beta \in \{0, 0.05, 0.1, 0.5, 1\}$. Configurations with $\beta=0.5$ and $\beta=1$ are not reported, as their KL-divergence term dominated the loss and prevented meaningful optimization of the evasion reward. We also omit the configuration with $\beta=0$, since the absence of KL regularization led to degenerate generations characterized by repeated-token outputs.

\begin{figure}
    \centering
    \includegraphics[width=0.9\linewidth]{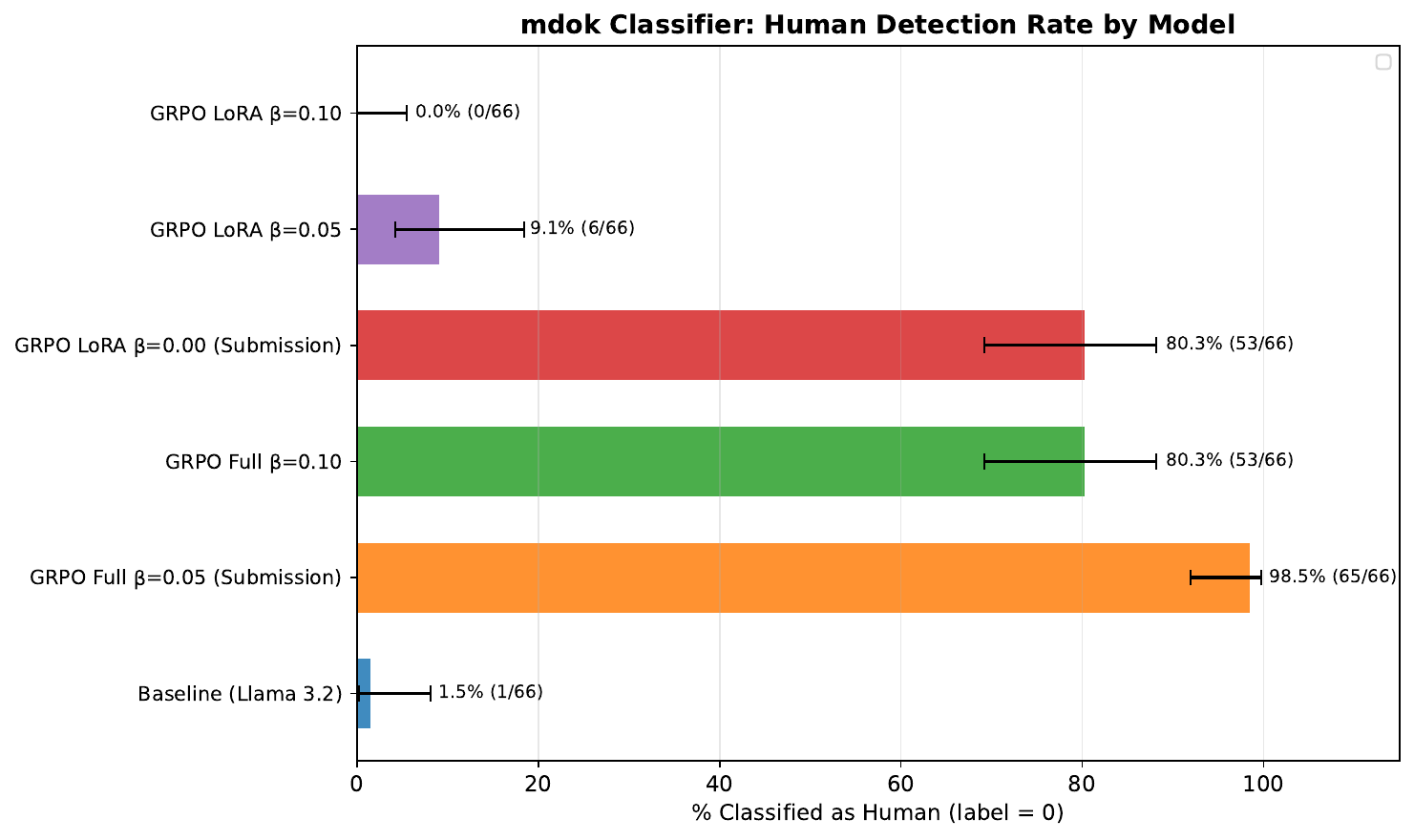}
    \caption{Generated text classified as human}
    \label{fig:confusion}
\end{figure}

Figure~\ref{fig:confusion} reports the percentage of generated texts classified as human by the surrogate \textit{mdok} detector for each evaluated configuration. The baseline \texttt{LLaMA~3.2} model achieved a human classification rate of only $1.5\%$ ($1/66$), indicating that its outputs were almost always identified as AI-generated. GRPO substantially improved evasion, but its effectiveness depended strongly on the fine-tuning strategy and the KL penalty. Full fine-tuning with $\beta=0.05$ achieved the strongest result, with $98.5\%$ of outputs classified as human ($65/66$). Increasing the penalty to $\beta=0.10$ reduced the rate to $80.3\%$ ($53/66$), suggesting that stronger regularization limits the policy movement required for detector evasion. 

The LoRA configurations followed a different pattern. Without KL regularization, LoRA achieved $80.3\%$ human classification ($53/66$), matching full fine-tuning with $\beta=0.10$. However, adding regularization sharply reduced evasion, reaching only $9.1\%$ ($6/66$) for $\beta=0.05$ and $0\%$ ($0/66$) for $\beta=0.10$. Overall, only three configurations exceeded the $50\%$ human-classification threshold: LoRA with $\beta=0$, full fine-tuning with $\beta=0.10$, and full fine-tuning with $\beta=0.05$. Among these, full fine-tuning with $\beta=0.05$ was the only configuration approaching near-complete surrogate evasion.

It is important to note that these results were obtained with the retry-based generation strategy described above. hen generation is performed in a one-shot setting, full fine-tuning with $\beta=0.05$ achieves a human-classification rate of $59.1\%$ ($39/66$), substantially lower than with retry-based selection. A similar drop is observed for the LoRA configuration with $\beta=0$, where only $30.3\%$ ($20/66$) of the generated texts are classified as human-written in the one-shot setting. These results indicate that the final evasion performance depends not only on the fine-tuned generator, but also on the inference-time selection strategy, which increases the probability of selecting outputs that better evade the surrogate detector.

\begin{figure}
    \centering
    \includegraphics[width=0.9\linewidth]{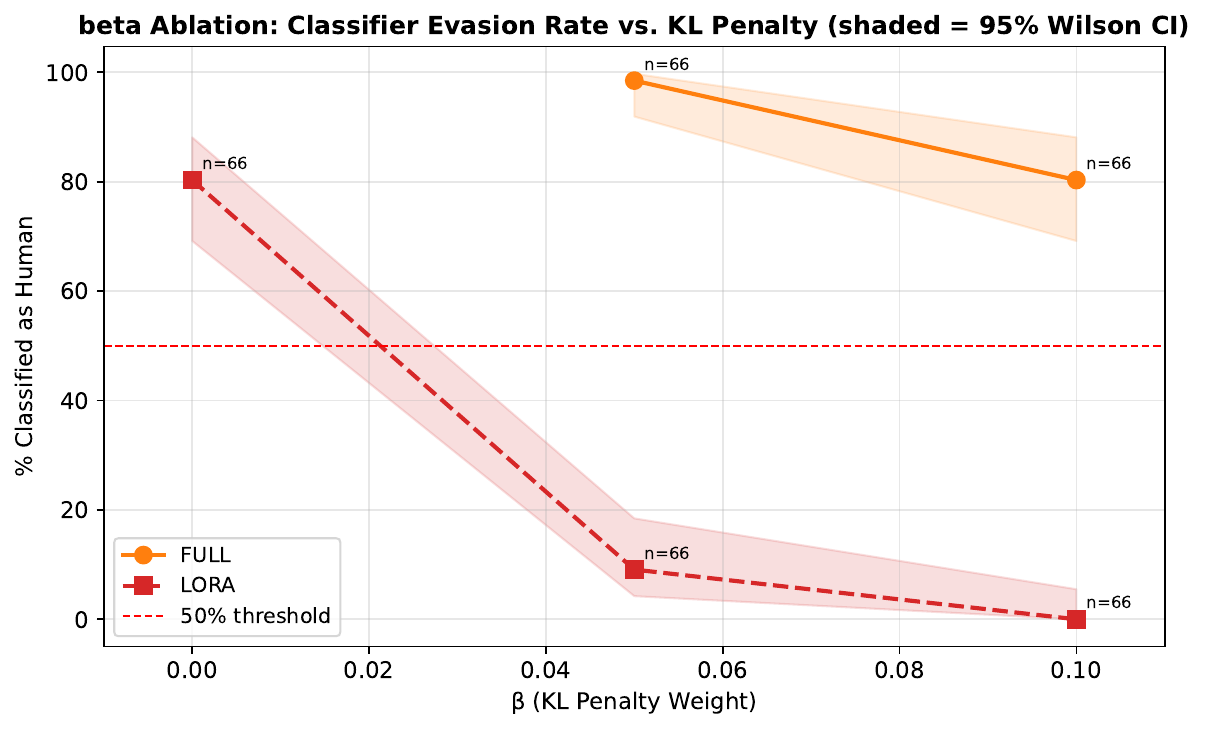}
    \caption{Beta vs. Confusion tradeoff}
    \label{fig:confusion_tradeoff}
\end{figure}

\Cref{fig:confusion_tradeoff} summarizes the same results as a function of the KL coefficient, making the regularization trade-off more explicit. Full fine-tuning remains effective under moderate KL regularization, whereas LoRA rapidly loses evasion capability as $\beta$ increases. This suggests that low-rank adapters are less able to simultaneously preserve proximity to the reference model and optimize the adversarial reward. The Wilson confidence intervals reflect the limited evaluation size ($n=66$), but they do not alter the qualitative trend. Taken together, these results indicate that a small, non-zero KL penalty is beneficial in the full fine-tuning setting because it prevents degenerate policy drift while still allowing sufficient capacity for adversarial adaptation.

\begin{figure}
    \centering
    \includegraphics[width=0.9\linewidth]{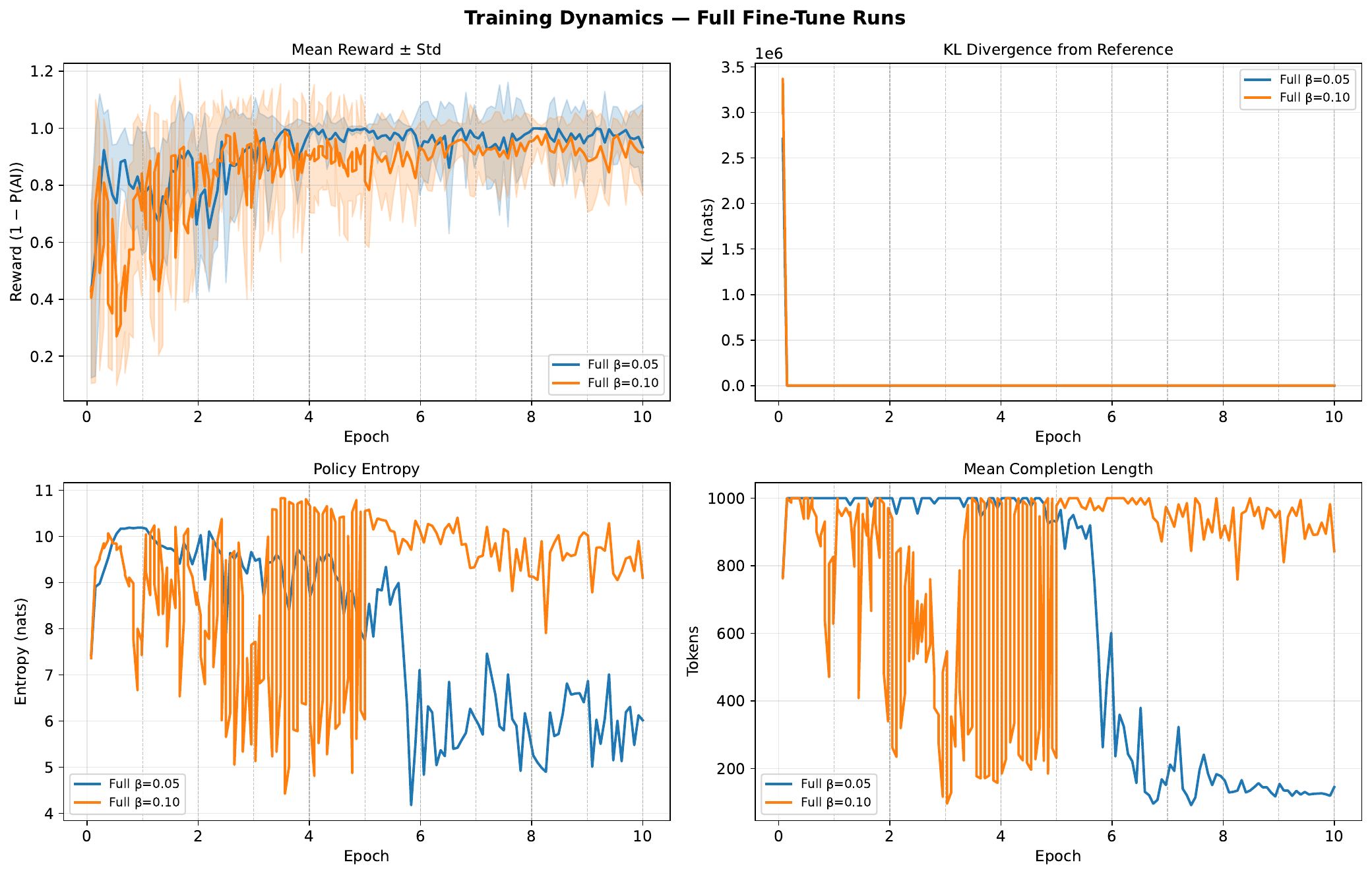}
    \caption{Training metrics full model}
    \label{fig:training_full}
\end{figure}

\begin{figure}
    \centering
    \includegraphics[width=0.9\linewidth]{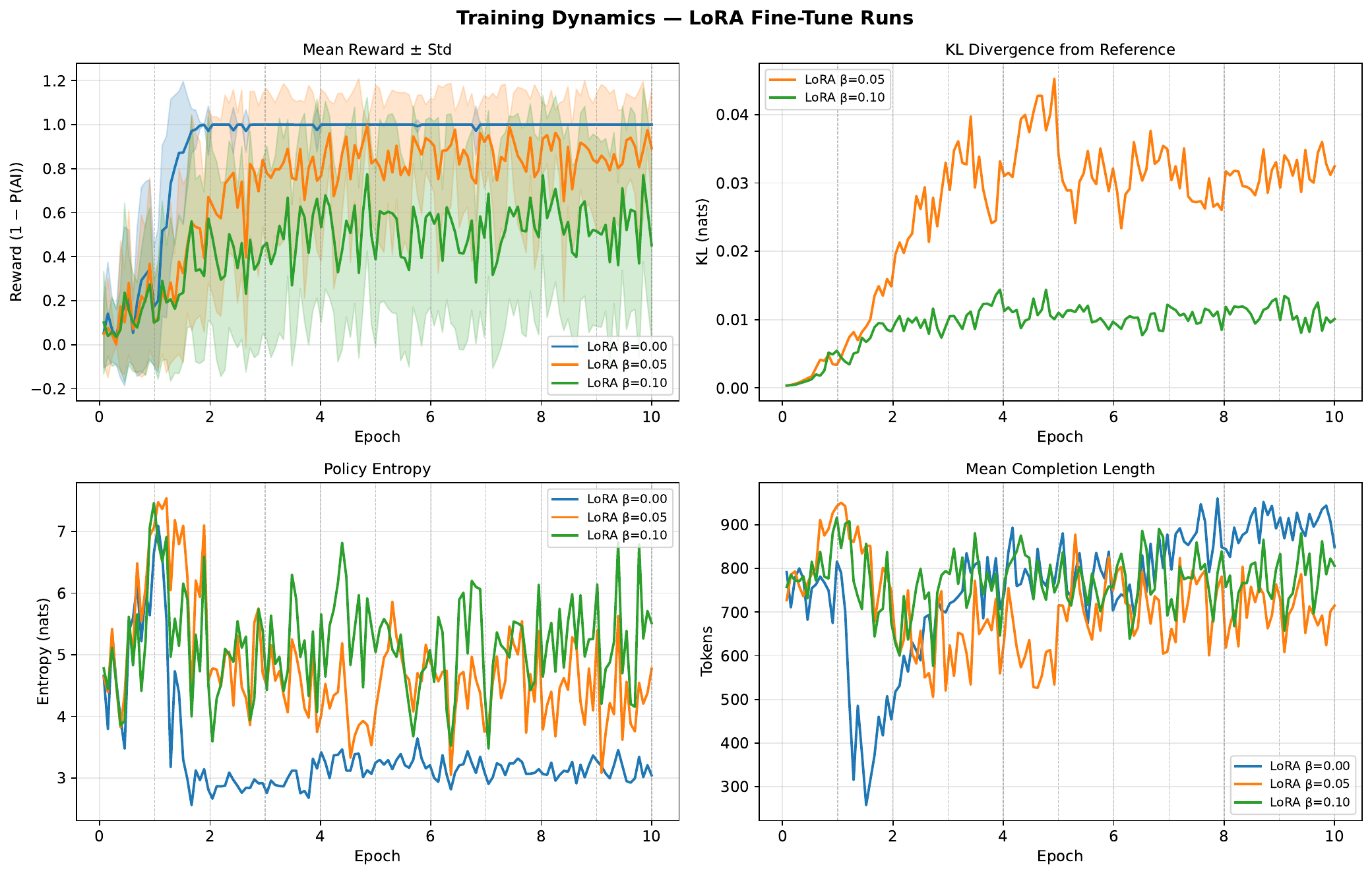}
    \caption{Training metrics LoRA model}
    \label{fig:training_lora}
\end{figure}

\Cref{fig:training_full} and \Cref{fig:training_lora} illustrate the training dynamics for the full and LoRA fine-tuning strategies, respectively. In the full fine-tuning setting, both configurations rapidly increase the reward during the first epochs, indicating that the model learns to reduce the surrogate detector's AI-generated probability. The run with $\beta=0.05$ achieves a slightly higher and more stable reward than $\beta=0.10$, which is consistent with its stronger evasion performance in \Cref{fig:confusion}. The KL divergence shows a large initial spike and then quickly drops by several orders of magnitude, suggesting that most of the policy shift occurs early in training. After this initial adaptation, the model continues optimizing within a more stable region of the parameter space.

The entropy and completion-length curves further reveal how full fine-tuning adapts the generation policy. For $\beta=0.05$, policy entropy decreases sharply after approximately five epochs, while the average completion length drops from near the maximum generation length to much shorter outputs. This suggests that the best-performing configuration converges toward a more specialized generation strategy that the surrogate detector tends to classify as human-written. In contrast, the $\beta=0.10$ configuration maintains higher entropy and longer completions throughout training, indicating that stronger KL regularization preserves more of the original generation behavior but also limits evasion effectiveness.
 
The LoRA runs exhibit a different pattern. The unregularized LoRA configuration ($\beta=0$) reaches a high reward early in training, but this is accompanied by rapid entropy collapse and large fluctuations in completion length, suggesting a less stable policy. When KL regularization is introduced the LoRA configurations maintain higher entropy and more stable completion lengths, but their rewards remain substantially lower, especially for $\beta=0.10$. Moreover, the KL divergence for the regularized LoRA runs remains small throughout training, indicating that the adapters induce only limited deviations from the reference policy. This helps explain the reduced evasion performance observed in \Cref{fig:confusion} and \Cref{fig:confusion_tradeoff}. Under KL regularization, LoRA appears unable to modify the generation policy enough to consistently evade the surrogate detector. 

Taken together, the training curves suggest that successful evasion requires a pronounced shift in generation behavior. Full fine-tuning provides enough capacity for such a shift, particularly when a small KL penalty prevents degenerate policy drift while still allowing adaptation. In contrast, LoRA offers insufficient adaptation capacity once proximity to the reference model is enforced. This supports the interpretation that detector evasion is not simply a matter of preserving fluent generation while lowering detector confidence. Rather, the model appears to discover specific generation regimes that exploit the surrogate detector.

\begin{figure}
    \centering
    \includegraphics[width=0.9\linewidth]{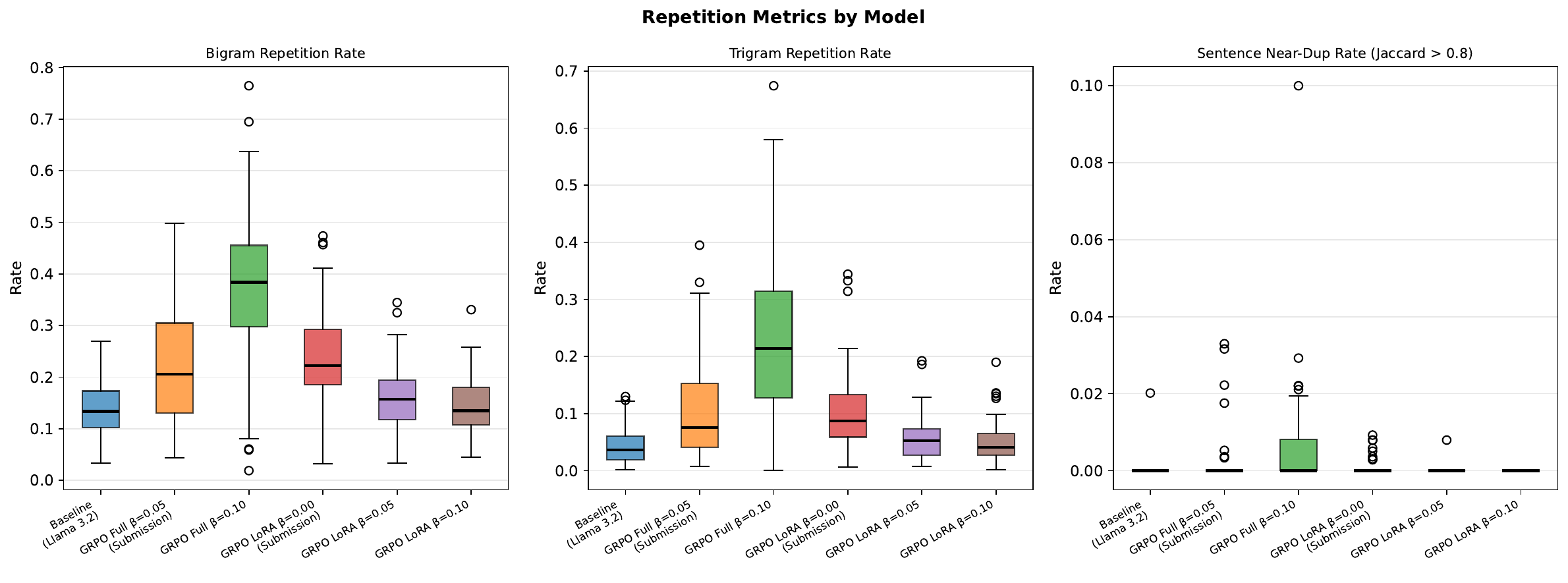}
    \caption{Repetitions within the generated text}
    \label{fig:repetitions}
\end{figure}

\Cref{fig:repetitions} analyzes repetition patterns in the generated texts using bigram repetition, trigram repetition, and sentence near-duplication metrics. The baseline model exhibits the lowest repetition levels overall, suggesting that the original instruction-tuned model produces more locally diverse outputs. In contrast, the full fine-tuned GRPO models introduce substantially higher repetition rates, particularly for $\beta=0.10$, which shows the highest median bigram and trigram repetition among all configurations. This indicates that full-policy adaptation can alter local lexical structure in ways that increase repeated $n$-gram patterns. Interestingly, this effect is stronger for $\beta=0.10$ than for the best-performing $\beta=0.05$ configuration, suggesting that higher repetition is not itself sufficient to explain surrogate evasion. 

The LoRA models generally preserved lower repetition levels than full fine-tuning, especially when KL regularization is applied. This is consistent with the earlier observation that regularized LoRA remains closer to the reference policy, but also achieves substantially lower evasion rates. Sentence-level near-deduplication remains close to zero across most configurations, with only isolated outliers. Thus, the observed degeneration mainly occurs at the local lexical level through repeated bigrams and trigrams, rather than through full sentence reuse. 
The contrast between the two full fine-tuning runs is particularly informative. Although $\beta=0.10$ produces more repetitive text, it achieves lower evasion than $\beta=0.05$. This suggests that \texttt{SHADE}'s strongest evasion performance is not simply caused by repetition, but by a broader shift in generation behavior.

\begin{figure}
    \centering
    \includegraphics[width=0.9\linewidth]{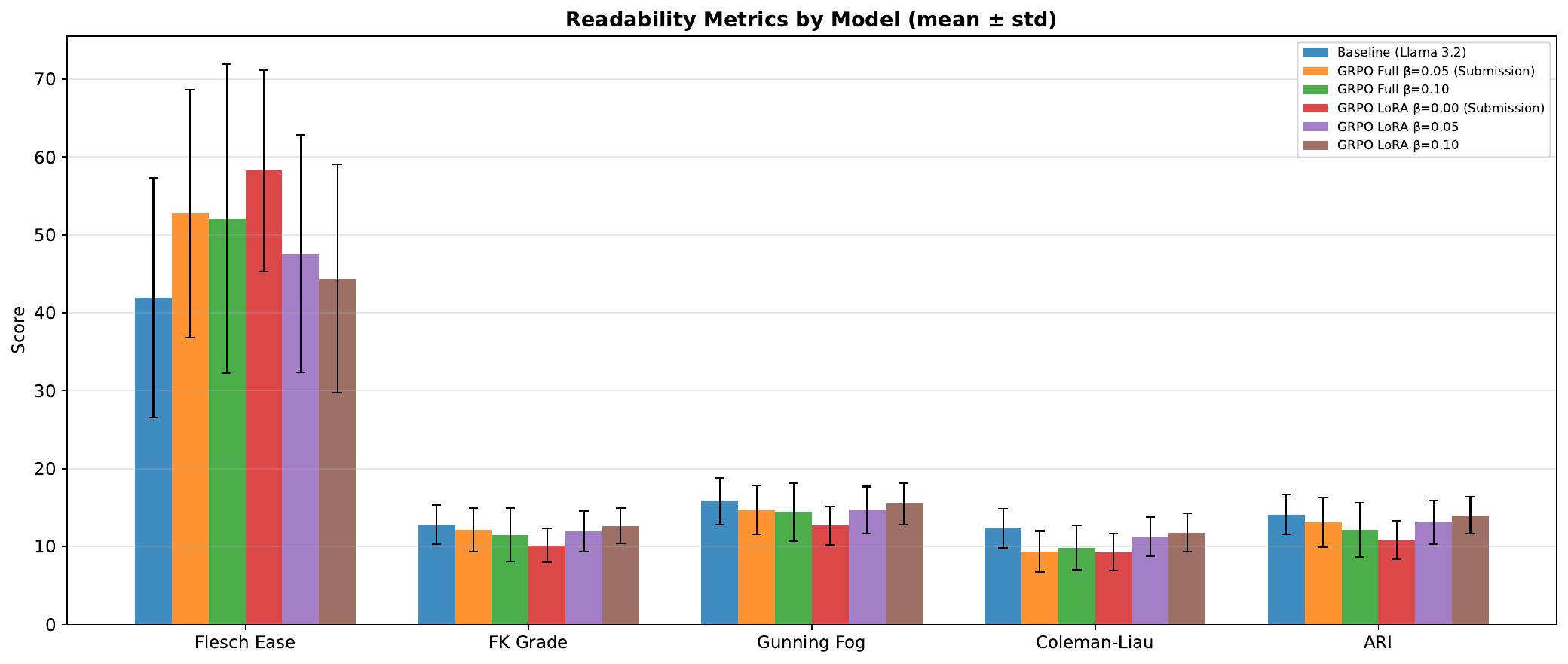}
    \caption{Generated text readability metrics}
    \label{fig:readability}
\end{figure}

\Cref{fig:readability} presents readability statistics across the evaluated models. Overall, the GRPO-based configurations tend to produce texts with higher Flesch Reading Ease scores and lower FK Grade values than the baseline, indicating a shift toward simpler and more accessible language. This pattern is especially visible for the high-evasion configurations. GRPO full fine-tuning with $\beta=0.05$ increases Flesch Ease from $41.9$ to $52.7$ while slightly reducing FK Grade from $12.8$ to $12.2$, and GRPO LoRA with $\beta=0$ achieves the highest Flesch Ease score and ($58.3$) and the lowest FK Grade ($10.1$). However, readability alone does not fully explain evasion performance. For instance, the LoRA $\beta=0$ configuration is the most readable but reaches only $80.3\%$ evasion, while full fine tuning with $\beta=0.05$ achieves $98.5\%$ evasion with less extreme readability changes. 

\Cref{tab:model_comparison} provides a broader view of these shifts. The best-performing configuration, GRPO full fine-tuning with $\beta=0.05$, produces shorter texts than the baselines ($374 \pm 136$ vs. $551 \pm 26$ words), with fewer sentences, lower type-token ratio, and lower hapax proportion. These changes suggest that detector evasion is associated with a reduction in lexical richness and structural complexity. At the same time, the comparison between full fine-tuning with $\beta=0.05$ and $\beta=0.10$ shows that excessive repetition is not sufficient for stronger evasion. The $\beta=0.10$ model has substantially higher bigram and trigram repetition rates, but lower evasion. Taken together, the readability and lexical metrics suggest that \texttt{SHADE} succeeds by moving the generation policy toward simpler, shorter, and less lexically diverse outputs, while avoiding the strongest forms of local degeneration. 

\begin{table}[]
    \caption{Comparison of human-likeness and linguistic statistics across baseline and GRPO-trained models.}
    \label{tab:model_comparison}
    \centering
    \small
    \scalebox{.75}{
    \begin{tabular}{lcccccccccc}
\hline
\textbf{Model} & \textbf{Cls Err} & \textbf{Words} & \textbf{Sentences}& \textbf{TTR} & \textbf{Hapax} & \textbf{Flesch Ease} & \textbf{FK Grade} & \textbf{Bigram Rep} & \textbf{Trigram Rep} \\
\hline
Baseline (LLaMA 3.2) 
& 1.5\% 
& $551 \pm 26$ 
& 22.4 
& 0.478 
& 0.339 
& 41.9 
& 12.8 
& 0.1376 
& 0.0426 \\

\textbf{GRPO Full $\beta=0.05$} 
& 98.5\%
& $374 \pm 136$ 
& 13.8 
& 0.414 
& 0.258 
& 52.7 
& 12.2 
& 0.2231 
& 0.1060 \\

GRPO Full $\beta=0.10$ 
& 80.3\%
& $434 \pm 159$ 
& 18.6 
& 0.329 
& 0.189 
& 52.1 
& 11.5 
& 0.3605 
& 0.2140 \\

\textbf{GRPO LoRA $\beta=0.00$} 
& 80.3\% 
& $559 \pm 22$ 
& 25.0 
& 0.377 
& 0.227 
& 58.3 
& 10.1 
& 0.2387 
& 0.1057 \\

GRPO LoRA $\beta=0.05$ 
& 9.1\% 
& $554 \pm 34$ 
& 23.5 
& 0.454 
& 0.311 
& 47.6 
& 12.0 
& 0.1579 
& 0.0550 \\

GRPO LoRA $\beta=0.10$ 
& 0.0\% 
& $557 \pm 23$ 
& 22.2 
& 0.473 
& 0.336 
& 44.4 
& 12.7 
& 0.1438 
& 0.0503 \\
\hline
\end{tabular}
    }
\end{table}

\section{Voight-Kampff Task Result\label{sec:final_results}}

In addition to the surrogate-detector evaluation, we also analyze the results provided by the Voight-Kampff competition organizers. Unlike the previous experiments, which evaluate generated texts against our mdok-based surrogate detector, the competition setting measures how closely each submission resembles source-like behavior under the official evaluation procedure. Therefore, these results provide a direct estimate of how well the proposed evasion strategy transfers from the surrogate detector to the unseen evaluation setting. 
Our two submitted runs, \textit{DArgk - Full fine-tuned $\beta=0.05$} and \textit{DArgk - LoRA $\beta=0$}, ranked in the middle of the ranking in positions sixth and seventh, respectively. The \textit{DArgk - Full fine-tuned $\beta=0.05$} submission obtained a mean score of $0.320$, with $337$ lower, $432$ equal, and $1811$ higher comparisons. The \textit{DArgk - LoRA $\beta=0$} submission obtained a slightly worse mean score of $0.338$, with $269$ lower, $402$ equal, and $1909$ higher comparisons. 

The gap between our surrogate results and the competition ranking highlights an important limitation of detector-specific reinforcement learning. While \texttt{SHADE} substantially reduces the confidence of the surrogate detector, this advantage does not fully transfer to the official evaluation setting. This suggests that the model may have learned generation patterns that exploit the surrogate detector in particular, rather than detector-invariant properties of human-written text. 
The \textit{DArgk -  Full fine-tuned $\beta=0.05$}  run slightly outperformed the submitted \textit{DArgk - LoRA $\beta=0$} run, which is consistent with our internal finding that full fine-tuning provides more effective adaptation than parameter-efficient tuning. However, the relatively small difference between the two submitted runs in the official ranking suggests that the advantages observed against the surrogate detector are attenuated when evaluated against unseen classifiers. 
Overall, the competition results indicate that \texttt{SHADE} is effective as a surrogate-targeted evasion method, but that improving cross-detector generalization remains a central challenge.

\section{Conclusion\label{sec:conclusions}}
This paper presented \texttt{SHADE}, a reinforcement learning framework for generating texts that evade AI-generated text detectors. Instead of relying on post-hoc perturbations, translation, or manually designed prompting strategies, \texttt{SHADE} formulates detector evasion as a policy optimization problem. Using a surrogate detector based on the PAN 2025 \textit{mdok} system, we fine-tuned an instruction-tuned LLaMA model with GRPO and optimized the generator to reduce the surrogate detector's confidence that the generated text was AI-authored. 
Overall, \texttt{SHADE} demonstrates both the potential and the risks of reinforcement learning for detector evasion. The approach shows that LLM generation policies can be optimized to exploit detector feedback, but also that high surrogate evasion does not guarantee robust transfer to unseen evaluation settings or human-like text quality. These findings highlight the need for more robust, adaptive, and transparent evaluation protocols for AI-generated text detection.

Several limitations of this study suggest directions for future work.
First, our reinforcement learning objective relies on a single surrogate detector, which may bias the generator toward detector-specific artifacts. Future work should reduce this risk by training against ensembles of heterogeneous surrogate detectors, including statistical, neural, and watermark-aware models, or by periodically updating the surrogate detector in an adaptive adversarial training loop. 
Second, the evaluation set used for the internal analysis is relatively small, which limits the statistical strength of some comparisons. This limitation could be addressed by evaluating the method on larger benchmark collections and by testing whether the observed trends hold across multiple domains and datasets. 
Third, the reward function focuses primarily on detector evasion and does not directly optimize for semantic fidelity, fluency, coherence, or human judgments of naturalness. Although the KL penalty partially constrains the model, future work should incorporate multi-objective rewards that jointly balance evasion, semantic preservation, readability, lexical diversity, and fluency, complemented by human evaluation to determine whether the generated outputs are genuinely perceived as human-written.
Finally, our analysis is limited to a specific base model, detector family, and shared-task setting. Future work should therefore investigate the transferability of learned evasion strategies across different LLMs, languages, domains, and unseen detector families.

\paragraph{Ethical considerations.} This work addresses AI-generated text detector evasion, a topic with clear dual-use implications. While the proposed method can help evaluate the robustness of current detectors, similar techniques could be misused to conceal AI authorship in contexts where originality, accountability or trust are important. We therefore frame \texttt{SHADE} as a robustness evaluation approach and evaluate it within the controlled setting of the Voight-Kampff shared task. 
Our analysis reports not only evasion performance but also limitations and side effects. 
These results emphasize that high detector evasion should not be interpreted as evidence of genuinely human-like writing. More broadly, the findings caution against relying on single-detector decisions in high-stakes settings such as education, scientific publishing, journalism or content moderation. Future work should emphasize multi-detector robustness evaluation, transparent uncertainty reporting, human oversight and safeguards against deceptive deployment.

\section{Artifacts\label{sec:artifacts}}
The following are the required artifacts to replicate the submission:
\begin{itemize}
    \item Source code for training and generating the submission, including pretrained weights for the mdok-roberta model: \href{https://github.com/knife982000/voight-kampff-2026}{https://github.com/knife982000/voight-kampff-2026}.
    \item Pretrained generative model Full fine-tuned weights $\beta=0.05$: \href{https://huggingface.co/jmrodri/Llama-3.2_voight-kampff_beta_005}{https://huggingface.co/jmrodri/Llama-3.2\_voight-kampff\_beta\_005}.
    \item Pretrained generative model LoRA weights $\beta=0$: \href{https://huggingface.co/jmrodri/Llama-3.2_voight-kampff_lora_beta_000}{https://huggingface.co/jmrodri/Llama-3.2\_voight-kampff\_lora\_beta\_000}.
    \item PAN 2025 Voight-Kampff AI Detection dataset: \href{https://zenodo.org/records/14962653}{https://zenodo.org/records/14962653}
    \item Eloquent 2024, 2025, and 2026 Voight-Kampff dataset: \href{https://huggingface.co/datasets/Eloquent/Voight-Kampff}{https://huggingface.co/datasets/Eloquent/Voight-Kampff}
\end{itemize}

\section{Declaration on Generative AI}
Generative AI tools were used solely to assist with grammar correction and language polishing. All conceptual content, analysis, and final decisions remain the authors' responsibility.

\paragraph{Acknowledgments}
We are grateful for the support of CLAAUDIA through AI-Cloud for providing the computational infrastructure. 
This research was funded in whole or in part by the Austrian Science Fund (FWF): \href{https://doi.org/10.55776/COE12}{1255776/COE12}.

\bibliography{references}
\end{document}